\documentclass[runningheads]{llncs}

\usepackage{graphicx}
\usepackage{amsmath} 
\usepackage{multirow}
\usepackage{multicol}
\usepackage{booktabs}
\usepackage{array}
\usepackage{bbding} 
\usepackage{fancyhdr}

\begin{document}
\title{A Multi-Branch Hybrid Transformer Network for Corneal Endothelial Cell Segmentation}
%
%

\author{Yinglin Zhang\inst{1,6}\and 
Risa Higashita\inst{1,5}$^\star$\and    
Huazhu Fu\inst{7} \and  
Yanwu Xu\inst{8} \and   
Yang Zhang\inst{1} \and     
Haofeng Liu\inst{1} \and    
Jian Zhang\inst{6} \and     
Jiang Liu\inst{1,2,3,4}$^\star$}    
%
\institute{Department of Computer Science and Engineering, Southern University of Science and Technology, Shenzhen 518055, China \and
Cixi Institute of Biomedical Engineering, Chinese Academy of Sciences, China \and
Guangdong Provincial Key Laboratory of Brain-inspired Intelligent Computation, Department of Computer Science and Engineering, Southern University of Science and Technology, Shenzhen 518055, China \and
Research Institute of Trustworthy Autonomous Systems, Southern University of Science and Technology, Shenzhen 518055, China \and
Tomey Corporation, Nagoya 451-0051, Japan \and
Global Big Data Technologies Centre, University of Technology Sydney, NSW, Australia \and
Inception Institute of Artificial Intelligence, UAE \and
Intelligent Healthcare Unit, Baidu, Beijing 100085, China\\
\email{k-chen@tomey.co.jp}}
\maketitle              
\begin{abstract}
Corneal endothelial cell segmentation plays a vital role in quantifying clinical indicators such as cell density, coefficient of variation, and hexagonality. However, the corneal endothelium's uneven reflection and the subject's tremor and movement cause blurred cell edges in the image, which is difficult to segment, and need more details and context information to release this problem. Due to the limited receptive field of local convolution and continuous downsampling, the existing deep learning segmentation methods cannot make full use of global context and miss many details. This paper proposes a Multi-Branch hybrid Transformer Network (MBT-Net) based on the transformer and body-edge branch. Firstly, We use the convolutional block to focus on local texture feature extraction and establish long-range dependencies over space, channel, and layer by the transformer and residual connection. Besides, We use the body-edge branch to promote local consistency and to provide edge position information. On the self-collected dataset TM-EM3000 and public Alisarine dataset, compared with other State-Of-The-Art (SOTA) methods, the proposed method achieves an improvement.

\end{abstract}

\section{Introduction}
Corneal endothelial cell abnormalities may be related to many corneal and systemic diseases. Quantifying corneal endothelial cell density, the coefficient of variation, and hexagonality have essential clinical significance \cite{al-fahdawi2018a}. Cell segmentation is a crucial step to quantify the above parameters. Nevertheless, manual segmentation is time-consuming, laborious, and unstable. Therefore, an accurate and fully automatic corneal endothelial cell segmentation method is essential to improve diagnosis efficiency and accuracy.

 \begin{figure}
\includegraphics[width=\textwidth]{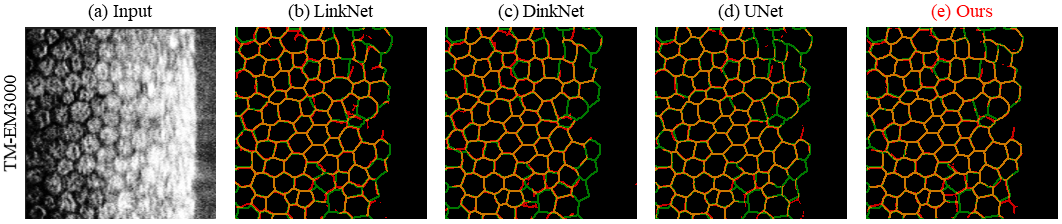}

\caption{Segmentation results on TM-EM3000.(a) is the input equalized corneal endothelium cell image.(b), (c), (d), (e) are the segmentation results of LinkNet, DinkNet, UNet, and our method. The red line represents the prediction result, and the green line represents ground truth, orange when the two overlap.} \label{fig1}

\vspace{-1.5em}
\end{figure}
The main challenge of accurate segmentation is the blurred cell edges, which are difficult to segment, as shown in Fig.\ref{fig1}, and needs more details and context information to release this problem. UNet \cite{ronneberger2015u-net} captures contextual semantic information through the contracting path and combines high-resolution features in the contracted path with upsampled output to achieve precise localization. UNet++ \cite{zhou2018unet++} optimizes it through a series of nested, dense skip connections to reduce the semantic gap between the encoder and decoder's feature maps. Fabij\'{a}nska \cite{fabijanska2018segmentation} first applied UNet to the task of corneal endothelial cell segmentation. Vigueras-Guillén et al. \cite{vigueras2019fully} applied the complete convolution method based on UNet and the sliding window version to the analysis of cell images obtained by SP-1P Topcon corneal endothelial microscope. Fu et al. \cite{fu2018multi} proposed a multi-conetxt deep network, by combining prior knowledge of regions of interest and clinical parameters. However, due to the limited receptive field of local convolution and continuous downsampling, they cannot make full use of the global context and still miss many details.

The transformer has been proved to be an effective method for establishing long-range dependencies. Vaswani et al. \cite{vaswani2017attention} proposed a transformer structure system for language translation tasks through a complete attention mechanism to establish the global dependence of input and output among time, space, and levels. Prajit et al. \cite{ramachandran2019stand-alone} explored the use of the transformer mechanism on visual classification tasks, replacing all spatial convolutional layers in ResNet with stand-alone self-attention layers. However, local self-attention will still lose part of the global information. Wang et al. \cite{wang2020axial} establish a stand-alone attention layer by using two decomposed axial attention blocks, to reduce the number of parameters and calculations, and allow performing attention in a larger or even global range.

Some previous works obtain better segmentation results by taking full advantage of edge information. Chen et al. \cite{00112016dcan} proposed the deep contour-aware network, using a multi-task learning framework to study the complementary information of gland objects and contours, which improves the discriminative capability of intermediate features. Chen et al.\cite{chen2016semantic} improved the network output by learning the reference edge map of CNN intermediate features. Ding et al. \cite{ding2019boundary-aware} proposed to use boundary as an additional semantic category to introduce boundary layout constraints and promote intra-class consistency through the boundary feature propagation module based on unidirectional acyclic graphs.

We need to preserve more local details and make full use of the global context. In this paper, we propose a Multi-Branch hybrid Transformer Network(MBT-Net). At first, we apply a hybrid residual transformer feature extraction module to give full play to the advantages of convolution block and transformer block in terms of local details and global semantics. Specifically, we use the convolutional block to focus on local texture feature extraction and establish long-range dependencies over space, channel, and layer by the transformer and residual connection. Besides, we define the corneal endothelial cell's segmentation task more entirely from the perspective of edge and body. Body-edge branches provide precise edge location information and promote local consistency. The experimental results show that the proposed method is superior to other state-of-the-art methods and has achieved better performance on two corneal endothelial datasets.

\section{Method}
\vspace{-1.5em}
 \begin{figure}
\includegraphics[width=\textwidth]{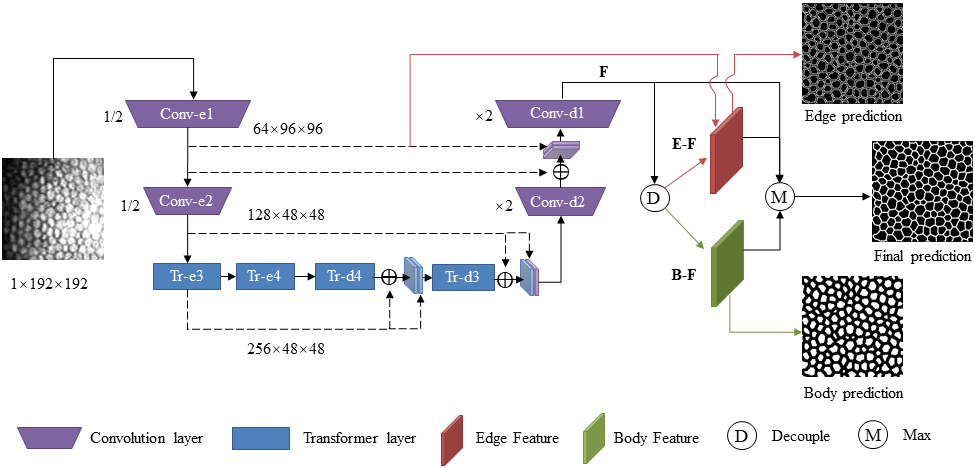}

\caption{The pipeline of multi-branch hybrid transformer network. Conv-e1, Conv-e2, Conv-d1 and Conv-d2 represent encoder and decoder layer based on convolution block. Tr-e3, Tr-e4, Tr-d3, and Tr-d4 are based on transformer blocks.} \label{fig2}
\vspace{-1.5em}
\end{figure}

In this paper, we propose the MBT-Net, as shown in Fig. 2. Firstly, the feature F of equalized corneal endothelium cell image is extracted by the hybrid residual transformer encoder-decoder module. Each convolution layer contains two basic residual blocks with a kernel size = 3×3. Each transformer layer contains two residual transformer blocks with kernel size = 1×48. Then, the feature is decoupled into two parts, body and edge. Also, the edge texture information from Conv-e1 is fused into the edge feature. Finally, we take the maximum response of edge feature E-F, body feature B-F, and feature F to obtain the fused feature to predict the final segmentation result. The training process of these three branches is explicitly supervised.

In this pipeline, the convolutional layer focuses on local texture feature extraction, which retains more details. The residual connection and transformer make full use of the feature map's global context information in a more extensive range of space, different channels, and layers. The edge perspective helps preserve boundary details, and the body perspective promotes local consistency. The low-resolution feature map of $d_{x+1}$  is refined by features from $e_x$ by concatenating and addition operation.

\subsection{Residual Transformer Block}

\begin{figure}
\vspace{-1.5em}
\begin{center}
\includegraphics[width=0.8\textwidth]{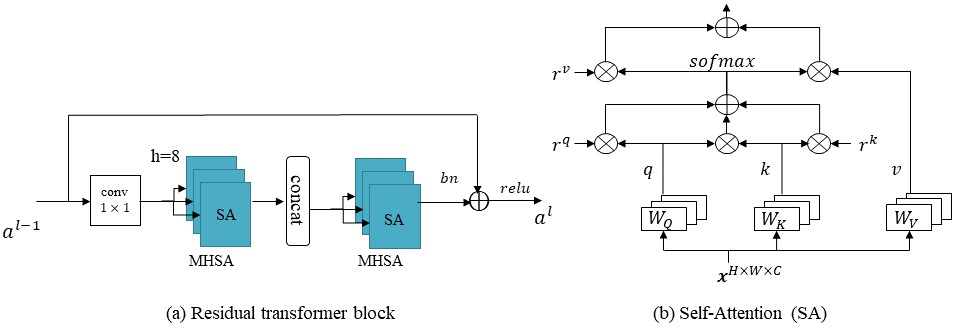}
\end{center}
\vspace{-1.5em}
\caption{The residual transformer block contains two 1×1 convolutions, a height-axial and a width-axial Multi-Head Self-Attention block (MHSA). MHSA compute axial self-attention (SA) with eight head. $r$, $W_Q$, $W_K$, and $W_V$ are learnable vectors, where $r$ related to the relative position} \label{fig3}

\vspace{-1.5em}
\end{figure}

The residual transformer block \cite{wang2020axial} is shown in Fig.\ref{fig3}, which contains two 1$\times$1 convolution to control the number of channels to be calculated and a height-axial and a width-axial Multi-Head Self-Attention block (MHSA), which significantly reduces the amount of calculation. This setting allows the transformer layer to consider the global spatial context in feature map size straightly. The axial Self-Attention(SA) module is defined as:

\vspace{-0.5em}
\begin{equation} \label{eq1}
y_{o} = \sum_{p \in N_{1 \times m(o)}} softmax_{p}(q_{o}^T k_{p} + q_{o}^{T}r_{p-o}^{q}+k_{p}^T r_{p-o}^k)(v_{p}+r_{p-o}^{v})
\end{equation}
\vspace{-0.5em}

For a given input feature map $x$, queries $q=W_{Q}x$, keys $k=W_{K}x$, values $v=W_{V}x$ are linear projections of feature map $x$, where $W_{Q}, W_{K}, W_{V}$ are learnable parameters. $r_{(p-o)}^{q}, r_{(p-o)}^{k}, r_{(p-o)}^{v}$ measure the compatibility from position $p$ to $o$ in query, key and value. They are also learnable paramters. The $softmax_{p}$ denotes a softmax function applied to all possible $p$ positions. $N_{1 \times m}(o)$ represents the local $1 \times m$ square region centered around location $o$, $y_{o}$ is the output at position $o$.

\vspace{-1em}
\begin{equation} \label{eq2}
    \begin{split}
    a^{l_{2}} &= a^{l_{1}} + \sum_{i=l_{1}}^{l_{2}-1}f(a^{i})
    \end{split}
\end{equation}
\vspace{-0.5em}

Besides, all the block used in encoder-decoder is in residual form, which can propagate input signal directly from any low layer to the high layer, optimizing information interaction \cite{he2016deep}, \cite{he2016identity}. Taking any two layers $l_{2}>l_{1}$ into consideration, the forward information propagation process is formulated as Eq.(\ref{eq2}).

\vspace{-1.0em}
\subsection{The Body, Edge, and Final Branches}

The information from the body and edge perspectives is combined to better define corneal endothelial cell segmentation. The body branch provides general shape and overall consistency information to promote local consistency, while the edge branch provides edge localization information to improve the segmentation accuracy of image details. 

We decouple the feature $F$ extracted by the hybrid residual transformer encoder-decoder module into $F_{body}=\phi(F)$ and $F_{edge} = F - F_{body}$, where $\phi$ is implemented by convolution layer. Also, the low level information from encoder Conv-e1 is fused into the edge feature, $F_{edge} = F_{edge} + \psi (F_{e1})$, where $\psi$ is dimension operation. Finally, the above three feature maps are fused into $F_{final}=\varphi(F, F_{edge}, F_{body})$ for final segmentation prediction, where $\varphi$ represents the maximum response. 

\begin{figure}
\vspace{-1.5em}
\begin{center}
\includegraphics[width=0.8\textwidth]{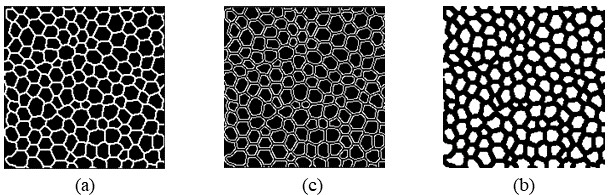}
\end{center}
\vspace{-1.5em}
\caption{Three kinds of masks on TM-EM3000.(a) The final prediction mask from the annotation of an expert, (b) The edge prediction mask extracted from (a) through the canny operator, (c) The body prediction mask by relaxing the edge of invert image of (a) with a Gaussian kernel.} \label{fig4}
\vspace{-2em}
\end{figure}

The training process of these three branches is explicitly supervised. The three masks used in training are shown in Fig.\ref{fig4}. The final prediction mask is the ground truth annotated by experts. The edge prediction mask is extracted from the final prediction mask by the canny operator. The body prediction mask is obtained by inverting the final prediction and then performing Gaussian blurring at the edges.

\vspace{-1.0em}
\subsection{Loss Function}

\begin{equation}\label{eq4}
    Loss = \lambda_{1}L_{b}(\hat{y}_{b}, y_{b}) + \lambda_{2}L_{e}(\hat{y}_{e}, y_{e}) + \lambda_{3}L_{f}(\hat{y}_{f}, y_{f})
\end{equation}

In this paper, we jointly optimize the body, edge, and final losses, as shown in Eq.(\ref{eq4}), where $\lambda_{1}, \lambda_{2}, \lambda_{3}$ are hyper parameters to adjust the weight of three different losses. As the final prediction is the output we finally use to compare with ground truth, we give it a higher weight than edge and body branch. In our experiment, we set $\lambda_{1}=0.5, \lambda_{2}=0.5, \lambda_{3}=1.2$. $y_{b}, y_{e}, y_{f}$ represent the ground truth of body, edge and final prediction respectively, and $\hat{y}_{b}, \hat{y}_{e}, \hat{y}_{f}$ are corresponding prediction from model. The binary cross entropy loss is used, as shown in Eq.(\ref{eq5}).

\begin{equation} \label{eq5}
    L = \frac{1}{N} \sum_{i}[y_{i}\ln \hat{y}_{i} + (1-y_{i})\ln (1-\hat{y}_{i})]
\end{equation}

Where N represents the total number of pixels, $y_{i}$ denotes target label for pixel $i$, $\hat{y}_{i}$ is the predicted probability.

\section{Experiments}

\subsection{Datasets and implementation Details }

\textbf{TM-EM3000} contains 184 images of corneal endothelium cell and its corresponding segmentation ground truth, with size = 266$\times$480, collected by specular microscope EM3000, Tomey, Japan. To reduce the interference of lesions and artifacts and build a data set with almost the same imaging quality, we select a patch with a size of 192$\times$192 from each image. This dataset is manually annotated and reviewed by three independent experts. We split it into the training set 155 patches, the validation set 10 patches, and the test set 19 patches.

\noindent \textbf{Alizarine Dataset} is collected by inverse phase-contrast microscope (CK 40, Olympus) at 200$\times$ magnification \cite{ruggeri2010a}. It consists of 30 images of corneal endothelium acquired from 30 porcine eyes and its corresponding segmentation ground truth, with image size = 768$\times$576, and mean area assessed per cornea = 0.54 ± 0.07 mm$^2$. Since each image in this dataset is only partly annotated, we select ten patches of size 192$\times$192 from each image to have 300 patches in total. And then split it into the training set 260 patches, validation set 40 patches.The training set and validation set do not overlap.

\noindent \textbf{Implementation Details}. We use the RMSprop optimization strategy during model training. The initial learning rate is 2e-4, epochs = 100 , batch size = 1. The learning rate optimization strategy is ReduceLROnPlateau, and the network input size is 192$\times$192. All the models are trained and tested with PyTorch on the platform of NVIDIA GeForce TITAN XP.

\subsection{Comparison with SOTA methods}

We compare performance of the proposed method with LinkNet \cite{chaurasia2017linknet}, DinkNet \cite{zhou2018d-linknet}, UNet \cite{ronneberger2015u-net}, UNet++\cite{zhou2018unet++} and TransUNet\cite{chen2021transunet} on \textbf{TM-EM3000} and \textbf{Alisarine} dataset. We use dice coefficient(DICE), F1 score(F1), sensitivity(SE), and specificity(SP)  as evaluation indicators, where DICE and F1 are most important because they are related to the overall performance.

\begin{table} [htb]

\caption{Quantitative evaluation of different methods. The proposed method achieves the best performance.}\label{tab1}
\vspace{-0.5em}
\begin{center}
\begin{tabular}{l l c c c| c c c c} 
\hline
\multirow{2}{*}{Model}&  \multicolumn{4}{c}{TM-EM3000}&   \multicolumn{4}{c}{Alisarine}\\
\cline{2-9}
&      DICE&    F1&     SE&     SP&     DICE&   F1&     SE&     SP\\
\hline
LinkNet34 \cite{chaurasia2017linknet}&      
0.711&  0.712&  0.719&  0.941&  0.766&  0.801&  0.805&  0.956  \\ 
DinkNet34 \cite{zhou2018d-linknet}&      
0.717&  0.718&  0.724&  0.944&  0.767&  0.805&  0.821&  0.953  \\ 
UNet \cite{ronneberger2015u-net}&           
0.730&  0.743&  0.763&  0.945&  0.775&  0.811&  0.814&  \textbf{0.960}  \\
UNet++ \cite{zhou2018unet++}&  
0.728&  0.739&  \textbf{0.775}&  0.938&  0.773&  0.811&  0.850&  0.947 \\
TransUNet \cite{chen2021transunet}&  
0.734&  0.742&  0.769&  0.941&  0.783&  0.821&  0.866&  0.948   \\

\textbf{Proposed}&
\textbf{0.747}&     \textbf{0.747}&     0.768&     \textbf{0.946}&     
\textbf{0.786}&     \textbf{0.821}&     \textbf{0.877}&    0.944   \\ 

\hline
\end{tabular}
\end{center}
\end{table}

As shown in Table \ref{tab1}, The proposed method has obtained the best overall performance on both TM-EM3000 and Alisarine data sets. On TM-EM3000, the DICE accuracy and F1 score of our approach are 0.747 and 0.747. On the Alisarine data set, the Dice accuracy and F1 score of our method are 0.786 and 0.821. UNet++\cite{zhou2018unet++} is modified from UNet, through a series of nested, dense skip connections to capture more semantic information. However, in general, there is no noticeable improvement observed in this experiment. TransUNet\cite{chen2021transunet} optimized the UNet by using the transformer layer to capture the global context in the encoder part, but our method has achieved better performance. It is mainly due to the following advantages. 1) Long-range dependencies are established through the transformer in both the encoder and decoder. 2) Performing transformer layer on the whole feature map, further reducing the loss of semantic information. 3) The body-edge branch encourages the network to learn more general features and provide edge localization information.

\begin{figure}
\includegraphics[width=\textwidth]{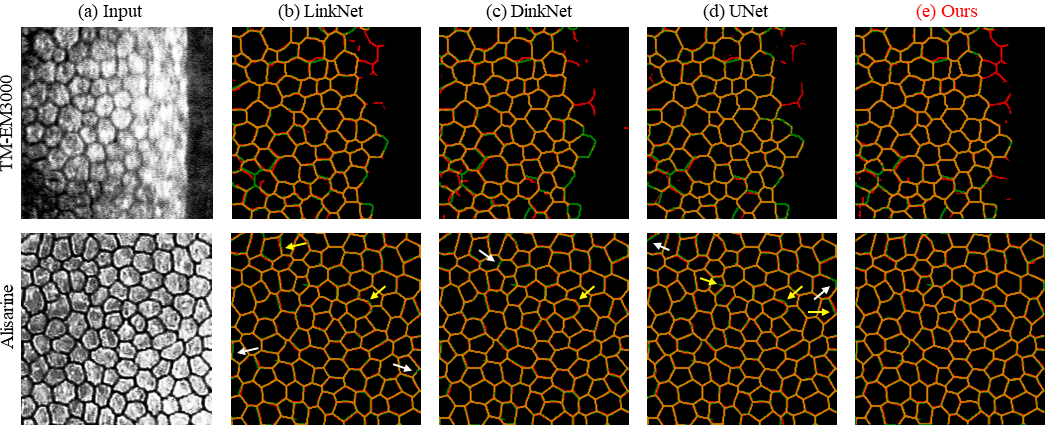}
\vspace{-1.5em}
\caption{Qualitative results on TM-EM3000 and the Alisarine Dataset.(a) is the input equalized corneal endothelium cell image. The red line represents the prediction result, and the green line represents ground truth, orange when the two overlap. The white arrow indicates the missing segmentation location, and the yellow arrow indicates the location where the segmentation result does not fit well with the ground truth} \label{fig5}

\end{figure}

As shown in Fig.\ref{fig5}, on the left side of the TM-EM3000 image, the cell boundary is clear. The segmentation performance of different methods is not much different. Nevertheless, on the right side with uneven illumination and the blur cell boundary, the proposed method achieves better segmentation results, closer to the ground truth, and in line with the real situation.

There is no extensive range of fuzzy area in the Alisarine image, and all methods obtained satisfied segmentation results. However, the baselines have varying degrees of loss in details, which lead to discontinuous cell edge segmentation as the white arrow indicated, and the segmentation results do not match the ground truth well as the yellow arrow indicated. The proposed method obtains better segmentation accuracy.

\subsection{Ablation Study}
The ablation experiment is conducted to explore the transformer's replacement design, as shown in Table \ref{tab2}. In this process, we gradually replace the encoder-decoder structure's convolution layer with the transformer from inside to outside. In the beginning, the model captures more semantic information, and the performance is improved.  Then, with the transformer replacing the shallow convolutional layer further, the model starts to lose local information, resulting in the decline of performance. Model 2-2-TR achieves the best balance between local details and global context.

\begin{table}[htb]
\caption{Ablation study on the replacement design of transformer. 0-0-TR means no transformer is used. 1-1-TR means e4, d4 is transformer layer. 2-2-TR means e3, e4, and d3, d4 is transformer layer. 3-3-TR means e2, e3, e4, and d2, d3, d4 is transformer layer. 4-4-TR means complete transformer structure.} \label{tab2}
    \begin{center}
    \begin{tabular}{c c c c c| c c c c} 
    \hline
    \multirow{2}{*}{Model}&  \multicolumn{4}{c}{TM-EM3000}&   \multicolumn{4}{c}{Alisarine}\\
    \cline{2-9}
    &      DICE&    F1&     SE&     SP&     DICE&   F1&     SE&     SP\\
    \hline
         0-0-TR&
         0.731& 0.737&  0.774&  0.937&      0.776&  0.813&  0.852&  0.948\\
         1-1-TR&    
         0.737&  0.746&  \textbf{0.778}&    0.941&  0.778&  0.816&  0.857&  \textbf{0.948}\\
         \textbf{2-2-TR}&    
         \textbf{0.747}&     \textbf{0.747}&     0.768&      \textbf{0.946}&
         \textbf{0.786}&     \textbf{0.821}&     \textbf{0.877}&    0.944 \\
         3-3-TR&    
         0.702&  0.714&  0.742&  0.935&    0.777&   0.812&  0.874&  0.940\\
         4-4-TR&    
         0.687&  0.707&  0.717&  0.940&     0.769&  0.802&  0.869&  0.936\\
    \hline
    \end{tabular}
    \end{center}
\vspace{-1em}
\end{table}

\begin{table}[htb]
\caption{Ablation study on transformer and body-edge branch on TM-EM3000.TR means transformer, and B-E means body-edge branch. When transformer is used, it means 2-2-TR.}
\label{tab3}
\begin{center}
\begin{tabular}{p{1.0cm} p{1.0cm} p{1.0cm} p{1.0cm} p{1.0cm} p{1.0cm}} 
\hline
\textbf{TR}    &\textbf{B-E}       
&\textbf{Dice}     &\textbf{F1}     &\textbf{SE}    &\textbf{SP}\\
\hline
\XSolidBrush    &\XSolidBrush       &0.720  &0.733  &0.746  &0.945 \\
\XSolidBrush    &\Checkmark         &0.731  &0.737  &0.774  &0.937 \\
\Checkmark      &\XSolidBrush       &0.736  &0.741  &\textbf{0.786}  &0.936 \\
\Checkmark      &\Checkmark         
&\textbf{0.747}  &\textbf{0.747}  &0.768  &\textbf{0.946}\\
\hline
\end{tabular}
\end{center}
\end{table}
\vspace{-1.5em}

We also study the influence of transformer and body-edge branch on performance on TM-EM3000 dataset, as shown in Table \ref{tab3}. When neither transformer nor body-edge branches are used, the DICE accuracy and F1 score on TM-EM3000 and Alisarine are 0.720 and 0.733, respectively. After adding the body-edge branch, the performance is improved to 0.731 and 0.737. When the transformer is used, the DICE accuracy and F1 score are 0.736 and 0.741. Using both the body-edge branch and transformer, we improve the performance by 2.7\% and 1.4\% in total to 0.747 and 0.747.

\vspace{-0.5em}
\section{Conclusion}
This paper proposes a multi-branch hybrid transformer network for corneal endothelial cell segmentation, which combines the convolution and transformer block's advantage and uses the body-edge branch to promote local consistency and provide edge localization information. Our method achieves superior performance to various state-of-the-art methods, especially in the fuzzy region. The ablation study shows that both the well-designed transformer replacement and body-edge branches contribute to improved performance.

%
%

%
%
%
\bibliographystyle{splncs04}
\bibliography{ref}

%


\end{document}